\documentclass[letterpaper]{article} 
\usepackage{aaai2026}  
\usepackage{times}  
\usepackage{helvet}  
\usepackage{courier}  
\usepackage[hyphens]{url}  
\usepackage{graphicx} 
\usepackage{natbib}  
\usepackage{multirow}
\usepackage{booktabs}
\usepackage{array}

\usepackage{caption} 
\usepackage{algorithm}

\usepackage{newfloat}
\usepackage{listings}
\usepackage{bm}

\usepackage{amsmath,amssymb,amsfonts}
\usepackage{textcomp}
\usepackage{amsthm}
\usepackage{pifont}
\usepackage{xcolor}

\usepackage{newfloat}
\usepackage{listings}

\usepackage{amsmath,amssymb,amsfonts}
\usepackage{textcomp}
\usepackage{amsthm}
\usepackage{pifont}
\usepackage{bm}                      

\usepackage[noend]{algpseudocode}  

\DeclareCaptionStyle{ruled}{labelfont=normalfont,labelsep=colon,strut=off} 
\floatstyle{ruled}
\newfloat{listing}{tb}{lst}{}
\floatname{listing}{Listing}
\title{Medverse: A Universal Model for Full-Resolution 3D Medical Image Segmentation, Transformation and Enhancement}
\author{
    Jiesi Hu\textsuperscript{\rm 1,2},
    Jianfeng Cao\textsuperscript{\rm 1},
    Yanwu Yang\textsuperscript{\rm 3,4},
    Chenfei Ye\textsuperscript{\rm 1},
    Yixuan Zhang\textsuperscript{\rm 1},
    Hanyang Peng\textsuperscript{\rm 2},
    Ting Ma\textsuperscript{\rm 1,2}
}

\affiliations{
    \textsuperscript{\rm 1}Harbin Institute of Technology at Shenzhen, Shenzhen, China\\
    \textsuperscript{\rm 2}Peng Cheng Laboratory, Shenzhen, China\\
    \textsuperscript{\rm 3}University Hospital Tübingen, Tübingen, Germany\\
    \textsuperscript{\rm 4}German Center for Mental Health, Germany
}

\usepackage{bibentry}

\begin{document}

\maketitle

\begin{abstract}
In-context learning (ICL) offers a promising paradigm for universal medical image analysis, enabling models to perform diverse image processing tasks without retraining. However, current ICL models for medical imaging remain limited in two critical aspects: they cannot simultaneously achieve high-fidelity predictions and global anatomical understanding, and there is no unified model trained across diverse medical imaging tasks (e.g., segmentation and enhancement) and anatomical regions. As a result, the full potential of ICL in medical imaging remains underexplored. Thus, we present \textbf{Medverse}, a universal ICL model for 3D medical imaging, trained on 22 datasets covering diverse tasks in universal image segmentation, transformation, and enhancement across multiple organs, imaging modalities, and clinical centers. Medverse employs a next-scale autoregressive in-context learning framework that progressively refines predictions from coarse to fine, generating consistent, full-resolution volumetric outputs and enabling multi-scale anatomical awareness. We further propose a blockwise cross-attention module that facilitates long-range interactions between context and target inputs while preserving computational efficiency through spatial sparsity. Medverse is extensively evaluated on a broad collection of held-out datasets covering previously unseen clinical centers, organs, species, and imaging modalities. Results demonstrate that Medverse substantially outperforms existing ICL baselines and establishes a novel paradigm for in-context learning. Our model are publicly available at \url{https://github.com/jiesihu/Medverse}.
\end{abstract}




\section{Introduction}
\label{sec:introduction}
Universal medical imaging models trained on large-scale, diverse datasets represent a significant step toward clinical AI deployment, exhibiting strong generalization across clinical centers and medical tasks~\cite{liu2023clip, butoi2023universeg, yang2024brainmass}.

Initially introduced in natural language processing~\cite{brown2020language}, in-context learning (ICL) has recently emerged as a unified paradigm in medical imaging~\cite{butoi2023universeg, gao2025show}. It conditions on a set of image–label pairs from other subjects, known as the context, to convey task information. By varying these context examples, a single model can perform diverse tasks such as segmentation, transformation, and enhancement, even including tasks that were not encountered during training~\cite{czolbe2023neuralizer}. Unlike traditional adaptation approaches, ICL adapts to distribution shifts and novel tasks without retraining~\cite{chan2022data, reddy2023mechanistic}, making it particularly suited for heterogeneous medical imaging scenarios.

\begin{figure}
\centering
\includegraphics[width=0.472\textwidth]{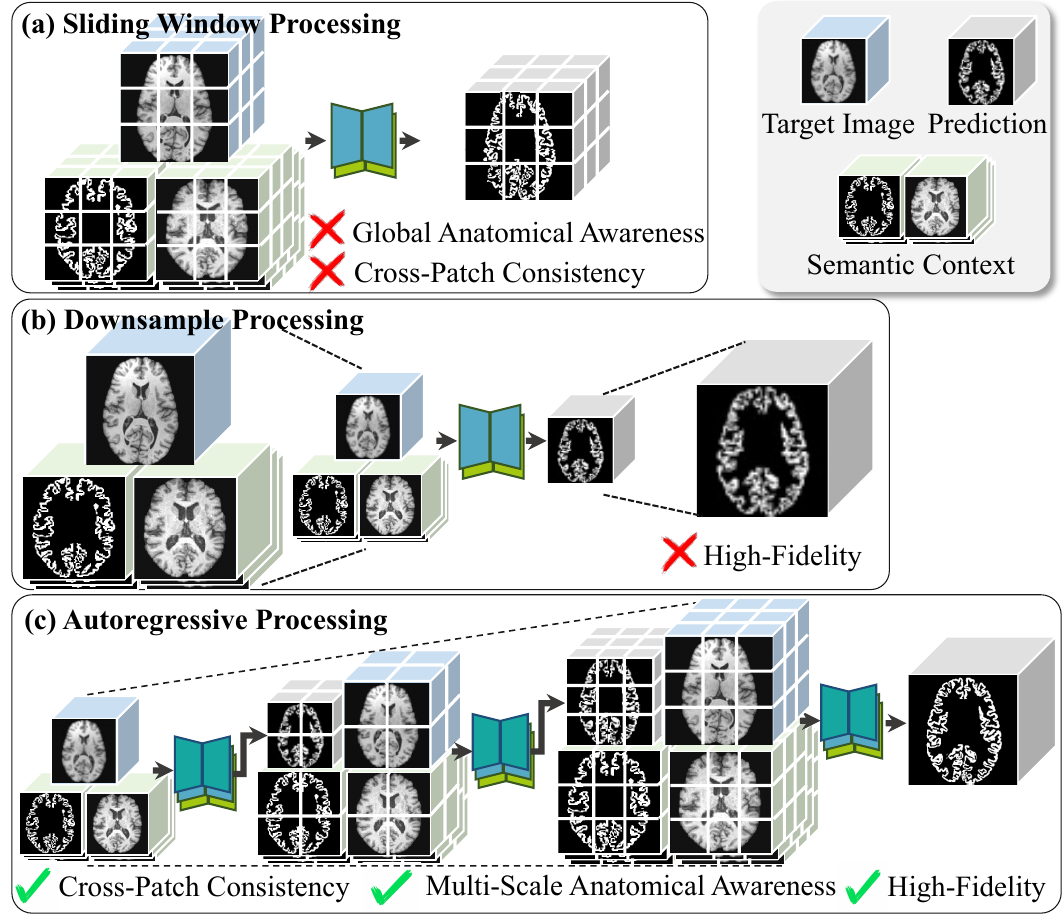}
\caption{
Illustration of different strategies for processing 3D medical images using ICL models. Due to the high resolution of volumetric data, direct end-to-end processing is often infeasible. The proposed next-scale autoregressive ICL framework progressively refines predictions from coarse to fine, producing full-resolution outputs.
}
\label{fig:intro_nsa}
\end{figure}

SegGPT~\cite{wang2023seggpt} and Painter~\cite{wang2023images} have demonstrated the effectiveness of ICL in natural image domains. However, their performance is suboptimal when applied to medical images~\cite{hu2025building}. Models such as UniverSeg~\cite{butoi2023universeg} and One-Prompt~\cite{wu2024one}, trained on medical imaging data, have shown promising results on 2D medical image segmentation tasks. Neuralizer~\cite{czolbe2023neuralizer} further extends ICL to a wider range of 2D tasks, including denoising and skull stripping. However, when applied to 3D data, they suffer from the loss of topological correlations and volumetric context. More recently, Neuroverse3D~\cite{hu2025building} has extended ICL to 3D neuroimaging by introducing adaptive context processing and architectural designs addressing the computational challenges of volumetric data. 

However, a major challenge is that current models remain constrained to operate at predetermined low spatial resolutions, which restricts their ability to capture fine-grained anatomical details and preserve the fidelity of output images. Attempts to increase output resolution via sliding-window strategies face the challenge of disrupting the anatomical continuity of context and target images, which impairs accurate understanding and the transmission of task-specific guidance from the context. This creates a dilemma for the development and practical use of ICL models in medical imaging. Another pressing issue is the absence of a unified 3D ICL model in medical imaging that is jointly trained across multiple organs and diverse types of image processing tasks, leaving the full potential of ICL in this domain underexplored.

To overcome the aforementioned issues, we present Medverse, a universal 3D medical image model trained on a diverse collection of datasets spanning multiple organs, imaging modalities, clinical centers, and task types, including universal segmentation, transformation, and enhancement. We introduce a \textbf{Next-Scale Autoregressive ICL framework (NA-ICL)} that adopts a coarse-to-fine prediction strategy and supports arbitrary input resolutions (Figure~\ref{fig:intro_nsa}). By integrating multi-scale anatomical cues from both context and target images, NA-ICL effectively balances global semantics with local details, improving prediction accuracy and producing consistent, full-resolution outputs while mitigating sliding-window artifacts. To enhance context perception under complex real-world conditions, we introduce a \textbf{Blockwise Cross-Attention Module (BAM)}, which enables long-range context-target interactions and alleviates spatial misalignment, while preserving computational and memory efficiency through spatial sparsity.

Our contributions are summarized as follows:
\begin{itemize}
    \item We introduce NA-ICL, a next-scale autoregressive in-context learning framework that enables the model to perform coarse-to-ﬁne prediction. This allows the model to leverage global semantics and local details, producing high-fidelity and spatially consistent, full-resolution outputs.
    \item We introduce BAM, which enables long-range context-target interactions while maintaining minimal memory and computational overhead through spatial sparsity.
    \item Trained on a diverse collection of datasets, Medverse is extensively evaluated on held-out data spanning unseen centers, organs, species, and modalities. It achieves strong generalization and consistently outperforms existing in-context learning baselines across segmentation, transformation, and enhancement tasks.
\end{itemize}




  
\section{Method}
\label{sec:method}

\begin{figure*}
\centering
\includegraphics[width=0.85\textwidth]{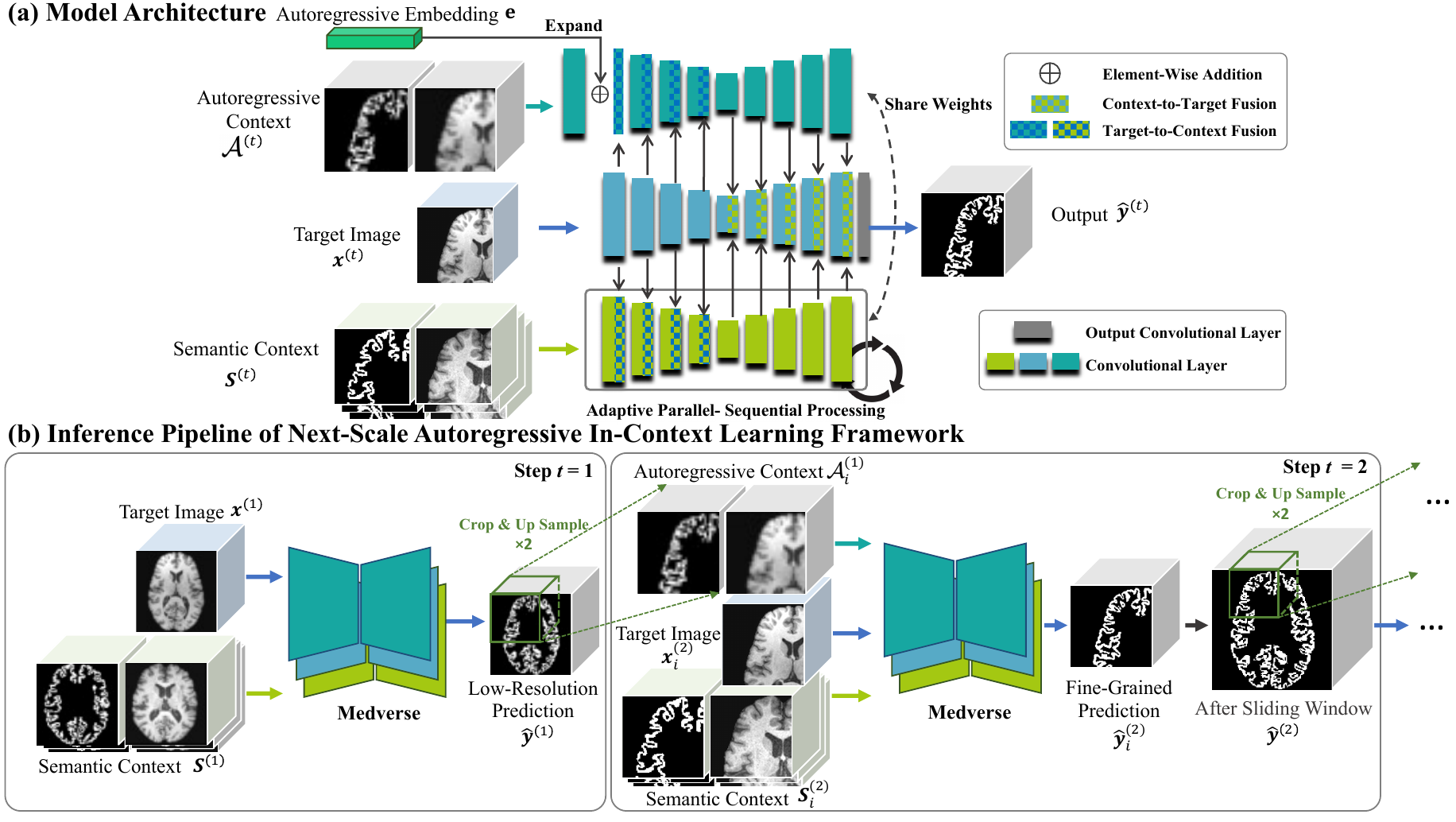}
\caption{Illustration of our model architecture and the inference pipeline of the next-scale autoregressive in-context learning framework.}
\label{fig:main_fig}
\end{figure*}

Under the ICL paradigm, Medverse achieves universality by learning task-specific guidance from context examples in the form of image-label pairs. It is capable of performing diverse tasks by leveraging the rich information embedded in these contexts.

\subsection{Next-Scale Autoregressive ICL}
Unlike conventional ICL models that utilize only context from other subjects, our NA-ICL framework provides the model with both image-label pairs from other subjects and the model's coarse predictions at lower resolutions, inspired by~\cite{tian2024visual}. This architectural design enables progressive processing from coarse to fine scales, offering benefits including the integration of multi-scale anatomical information, improved consistency across patches, and high-fidelity outputs.

\vspace{0.4em}
\noindent\textbf{Network Architecture.}
As shown in Figure~\ref{fig:main_fig} (a), the proposed model comprises three 3D U-Net branches that, from top to bottom, process the autoregressive context, the target image, and the semantic context. The target image branch interacts with both context branches through fusion modules to facilitate the context-target feature interaction.

The autoregressive context comprises the target image and its associated prediction from a lower-resolution step, and is treated as a specialized type of context. The semantic context consists of a set of image-label pairs drawn from other subjects, providing conventional semantic guidance for the task. To promote parameter efficiency and shared contextual understanding, the autoregressive context branch shares weights with the semantic context branch. To differentiate the features from the autoregressive context, we introduce a learnable autoregressive embedding $\bm{e} \in \mathbb{R}^{C}$, which is added to the output of the first layer of the autoregressive context branch as follows:
\begin{equation}
    \bm{f}^{'}_{1, \text{AR}} = \bm{f}_{1, \text{AR}} + \operatorname{Expand}(\bm{e}),
\end{equation}
where $\bm{f}_{1, \text{AR}} \in \mathbb{R}^{C \times H \times W \times D}$ denotes the feature map from the first layer and $\operatorname{Expand}(\cdot)$ expands the embedding across spatial dimensions.

Let the semantic context be defined as $S^{(t)}$, which contains a number of image-label examples. $t$ refers to the corresponding autoregressive step, which will be detailed later. Given the target image $\bm{x}^{(t)}$ and the autoregressive context from the previous step, $\mathcal{A}^{(t-1)} = (\bm{x}^{(t-1)}, \hat{\bm{y}}^{(t-1)})$, the model prediction at step $t$ is computed as
\begin{equation}
\hat{\bm{y}}^{(t)} = F(\bm{x}^{(t)},\, S^{(t)},\, \mathcal{A}^{(t-1)}),
\end{equation}
where $F(\cdot)$ denotes the prediction function of Medverse. To efficiently handle a large number of semantic context examples, we incorporate the adaptive parallel-sequential context processing module introduced in~\cite{hu2025building}, which enables scalable context processing with minimal memory requirements.

\vspace{0.4em}
\noindent\textbf{Autoregressive Inference Pipeline.}  
Given an input volume of size $(H,W,D)$, the number of autoregressive steps is
\begin{equation}
T=\bigl\lceil\log_{2}\bigl(\frac{\max\{H,W,D\}}{I}\bigr)\bigr\rceil+1,
\end{equation}
where $I\times I\times I$ denotes the model’s input patch size.
For step $t\in\{1,\dots,T\}$, we downsample the input images to
\begin{equation}
\bigl(H^{(t)},W^{(t)},D^{(t)}\bigr)=\bigl\lceil \frac{H}{2^{\,T-t}}\bigr\rceil,\;
\bigl\lceil \frac{W}{2^{\,T-t}}\bigr\rceil,\;
\bigl\lceil \frac{D}{2^{\,T-t}}\bigr\rceil,
\end{equation}
so that the resolution is doubled from level $t$ to $t+1$. We assume that the context and target images share the same spatial dimensions and undergo identical resizing operations. After the final step $t=T$, the model outputs a prediction at the original image resolution.

Figure~\ref{fig:main_fig} (b) illustrates the autoregressive inference pipeline. At step $t=1$, the model predicts $\hat{\bm{y}}^{(1)} = F(\bm{x}^{(1)},\, S^{(1)},\,\mathcal{A}^{(0)})$ on the low-resolution input image, capturing global anatomical context, with $\mathcal{A}^{(0)}$ initialized as an empty set. The autoregressive context $\mathcal{A}^{(1)} = (\bm{x}^{(1)}, \hat{\bm{y}}^{(1)})$ is then passed to step $t=2$. At $t=2$, the increased resolution requires the target image $\bm{x}^{(2)}$ to be split into $I \times I \times I$ patches, which are processed with a sliding window to capture finer anatomical detail. The prediction for each patch $\bm{x}^{(2)}_i$ is then given by $\hat{\bm{y}}^{(2)}_i = F(\bm{x}^{(2)}_i,\, S^{(2)}_i,\, \mathcal{A}^{(1)}_i)$. $S^{(2)}_i$ denotes the semantic context cropped at the same spatial location as $\bm{x}^{(2)}_i$ to ensure patch-wise alignment. To propagate global information from coarser scales, the corresponding region in $\mathcal{A}^{(1)}$ of size $\frac{I}{2} \times \frac{I}{2} \times \frac{I}{2}$ is extracted, upsampled to $I \times I \times I$, and used as $\mathcal{A}^{(1)}_i$. The same procedure is repeated for $t \ge 2$, with the number of sliding windows increasing as the resolution increases. A step-by-step pseudocode is provided in the supplementary material.

\begin{figure*}
\centering
\includegraphics[width=0.99\textwidth]{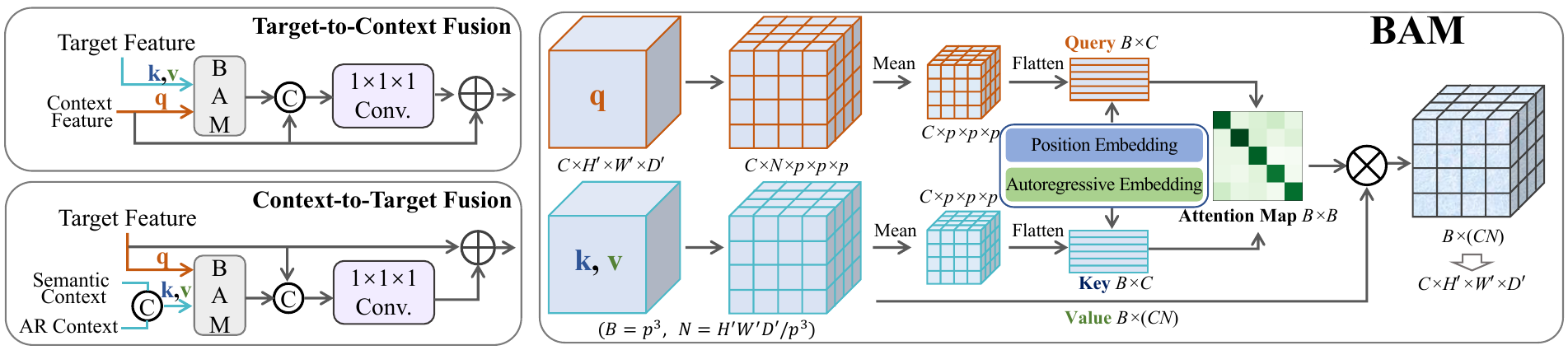}
\caption{Illustration of fusion modules and the blockwise cross-attention module.}
\label{fig:bam}
\end{figure*}

\subsection{Context-Target Fusion Module}  
Feature interaction between different branches in the network is achieved through fusion modules, enabling the target branch to effectively incorporate contextual information. As shown in Figure~\ref{fig:main_fig} (a), we perform target-to-context fusion within the encoder, where features from the target branch are passed to the context branch, and context-to-target fusion within the decoder, where features from the context branch are passed back to the target branch. The use of adaptive parallel-sequential context processing necessitates this order of interaction.

The details of the fusion module are illustrated in Figure~\ref{fig:bam}.
The target-to-context module employs the context feature map as the query, with the target feature map serving as the source of keys and values.
BAM transfers target information to the context, after which the output is concatenated with the original context, compressed by a $1\times1\times1$ convolution, and added back through a residual shortcut to produce a target-aware context representation.

The context-to-target fusion module concatenates semantic and autoregressive context features, using the combined representation as the source of keys and values, with the target features serving as the query. BAM injects both semantic and autoregressive contextual guidance into the target representation. The resulting features are then similarly passed through concatenation, convolution, and a residual shortcut to produce a context-aware target representation.


\vspace{0.4em}\noindent\textbf{Blockwise Cross-Attention Module.}  
Since there is not necessarily strict spatial alignment between the context and target objects, long-range interactions are crucial for effective feature fusion. We propose BAM, which enables efficient long-range feature aggregation with spatial sparsity.

We denote the query feature map by
$\bm{X}_{q}\!\in\!\mathbb{R}^{C\times H'\times W'\times D'}$, and
the key–value map by
$\bm{X}_{kv}\!\in\!\mathbb{R}^{C\times H'\times W'\times D'}$.
The volume is partitioned into
$p\times p\times p$ non-overlapping blocks,
so that each block covers
$h=\frac{H'}{p}$,
$w=\frac{W'}{p}$,
and
$d=\frac{D'}{p}$ voxels.
Let $N = hwd$ denote the number of voxels per block.
After partitioning, we obtain the blockwise features
$\bm{X}_{q}^{\text{Blockwise}},\, \bm{X}_{kv}^{\text{Blockwise}} \in \mathbb{R}^{C \times N \times p \times p \times p}$.
The query and key representations for each block are then computed by mean pooling over the $N$ voxels within the block:
\begin{align}
\bm{Q}' &= \frac{1}{N} \sum_{n=1}^{N} \bm{X}_{q}^{\text{Blockwise}}[:, n, :, :,:] \in \mathbb{R}^{C\times p\times p\times p}, \\
\bm{K}' &= \frac{1}{N} \sum_{n=1}^{N} \bm{X}_{kv}^{\text{Blockwise}}[:, n, :, :, :] \in \mathbb{R}^{C\times p\times p\times p}.
\end{align}

To match the input format required by the attention mechanism, we transpose and reshape $\bm{Q}'$ and $\bm{K}'$ to obtain
$\bm{Q}, \bm{K} \in \mathbb{R}^{B \times C}$, where $B = p^{3}$ denotes the total number of blocks.
Then, the queries and keys are linearly projected into a $m$-dimensional space via
learnable matrices $\bm{W}_Q, \bm{W}_K \in \mathbb{R}^{C \times m}$.
To incorporate spatial priors, we add a 3D sine–cosine positional embedding
$\bm{P} \in \mathbb{R}^{B \times m}$ to both branches.
In addition, for blocks originating from the autoregressive context,
we further add a learnable autoregressive embedding
$\bm{R} \in \mathbb{R}^{m}$,
which acts as a marker for features from the autoregressive branch and is broadcast across all blocks to shape $B \times m$.
The resulting transformed features are given by
\begin{align}
\widehat{\bm{Q}} &= \bm{Q} \bm{W}_Q + \bm{P} + \bm{1}_{\text{AR}} \cdot \bm{R}, \\
\widehat{\bm{K}} &= \bm{K} \bm{W}_K + \bm{P} + \bm{1}_{\text{AR}} \cdot \bm{R},
\end{align}
where $\bm{1}_{\text{AR}}$ is an indicator that equals 1 if the block comes
from the autoregressive context and 0 otherwise. The block-level attention weights are as follows:
\begin{align}
\bm{A}=
\operatorname{Softmax}\!\Bigl(
\frac{\widehat{\bm{Q}}\widehat{\bm{K}}^{\!\top}}{\sqrt{m}}
\Bigr)\in\mathbb{R}^{B\times B}.
\end{align}

To preserve spatial granularity, the value features are obtained directly from the unpooled input as $\bm{V}' = \bm{X}_{kv}^{\text{Blockwise}} \in \mathbb{R}^{C \times N \times p \times p \times p}$. They are then transposed and reshaped to match the input format for attention dot-product computation, yielding $\bm{V} \in \mathbb{R}^{B \times (C N)}$. The value matrix $\bm{V}$ is multiplied by the attention weights to obtain
\[
\bar{\bm{Y}} = \bm{A}\,\bm{V}
      \in \mathbb{R}^{B \times (C N)}.
\]
The result is then reshaped back to the original spatial layout,
producing the final BAM output:
\[
\widehat{\bm{Y}}
  = \operatorname{Reshape}\bigl(\bar{\bm{Y}}\bigr)
  \in \mathbb{R}^{C \times H' \times W'\times D'}.
\]

The computational complexity is thereby reduced to $\mathcal{O}(B^{2})$, in contrast to the original $\mathcal{O}\!\bigl((H'W'D')^{2}\bigr)$, while still preserving long-range interactions and retaining fine-grained value information. $B = p^3$ is a user-defined constant that remains fixed across U-Net stages, further ensuring stable computational cost. Compared to the direct concatenation-based fusion used in~\cite{butoi2023universeg,czolbe2023neuralizer, hu2025building}, this cross-attention fusion approach is better suited for handling spatial misalignment between the context and target objects.

\subsection{Loss Function}
For fair comparison, we adopt the same loss functions as those proposed in~\cite{hu2025building}. Specifically, a modified $\mathcal{L}_{\text{smooth}-L_{1}}$ loss is used for segmentation tasks, while for image enhancement and transformation tasks, the $\mathcal{L}_{\text{smooth}-L_{1}}$ loss is applied to both image intensity and intensity differences.

\begin{table*}[htbp]
\centering
\scriptsize
\setlength{\tabcolsep}{1.5pt}
\begin{tabular}{l|c|cccccc|ccc|c|c|c}
\toprule
\multirow{2}{*}{\textbf{Methods}}  & 
\multirow{2}{*}{\renewcommand{\arraystretch}{0.8}\begin{tabular}[x]{@{}c@{}}\textbf{Fine-}\\ \textbf{Tuning}\\\textbf{Free}\end{tabular}} & 
\multicolumn{6}{c|}{\textbf{\textit{Unseen Center}}} &  \multicolumn{3}{c|}{\textbf{\textit{Unseen Organ}}} &  \multicolumn{1}{c|}{\textbf{\textit{Unseen Species}}} & \multicolumn{1}{c|}{\textbf{\textit{Unseen Modality}}} & \multirow{2}{*}{\textbf{Average}}  \\
 &  & 
\renewcommand{\arraystretch}{0.8}\begin{tabular}[x]{@{}c@{}}Cerebral\\Cortex\end{tabular}& 
Hippocampus & 
Thalamus & 
Liver & 
Spleen & 
\renewcommand{\arraystretch}{0.8}\begin{tabular}[x]{@{}c@{}}Kidney\\Left\end{tabular}& 
\renewcommand{\arraystretch}{0.8}\begin{tabular}[x]{@{}c@{}}Maxillary\\Sinus\end{tabular}& 
\renewcommand{\arraystretch}{0.8}\begin{tabular}[x]{@{}c@{}}Nasal\\Cavity\end{tabular}& 
\renewcommand{\arraystretch}{0.8}\begin{tabular}[x]{@{}c@{}}Nasal\\Pharynx\end{tabular}& 
Mice Lung & 
\renewcommand{\arraystretch}{0.8}\begin{tabular}[x]{@{}c@{}}PET Lateral\\Ventricle\end{tabular}& \\
\midrule
\multicolumn{14}{l}{\textit{Fully Supervised Task-Specific Models (Upper Bound)}} \\
nnUNet  & \ding{55} & 90.30 &	90.99 &	93.89 &	98.46 &	96.60 &	96.06 &	94.07 &	91.63 &	94.63 &	94.49 &	84.26 &	93.22 \\
3D-Unet & \ding{55} & 88.55	 &89.73 &	92.88 &	96.41 &	96.52 &	91.10 &	90.13 &	89.02 &	92.64 &	94.21 &	82.35 &	91.23 \\
Swin-UNETR   & \ding{55} & 89.78 &	89.38 &	92.92 &	96.49 &	94.45 &	94.88 &	94.79 &	89.08 &	93.65 &	93.21 &	82.11 &	91.88\\
\midrule
\multicolumn{14}{l}{\textit{Few-Shot Task-Specific Models}} \\
3D-Unet      & \ding{55} & 87.90 &	86.66 &	90.56 &	94.95 &	81.74 &	81.29 &	86.77 &	86.99 &	90.05 &	91.89 &	75.95 &	86.80 \\
Swin-UNETR   & \ding{55} & 87.62 &	86.30 &	91.15 &	94.66 &	88.64 &	87.82 &	87.99 &	84.96 &	89.46 &	91.40 &	74.40 &	87.67 \\
\midrule
\multicolumn{14}{l}{\textit{ICL Models}} \\
SegGPT       & \ding{51} & 45.38 & 28.41 & 19.56 & 68.07 & 39.02 & 36.15 & 46.35 & 52.79 & 37.25 & 43.30 & 42.22 & 41.68 \\
Neuralizer   & \ding{51} & 69.20 & 57.49 & 45.11 & 73.54 & 52.12 & 62.71 & 75.77 & 64.79 & 73.65 & 70.48 & 51.83 & 63.34 \\
UniverSeg    & \ding{51} & 68.79 & 59.90 & 47.57 & 81.10 & 57.79 & 56.76 & 80.12 & 75.78 & 72.64 & 65.77 & 48.90 & 65.01 \\
SegGPT*       & \ding{51} & 50.83&	34.30&	50.47&	79.12&	57.96&	69.44&	64.68&	31.86&	56.38&	72.33&	42.54&	55.45\\
Neuralizer*   & \ding{51} & 76.96&	65.70&	82.79&	59.45&	62.69&	71.58&	83.64&	66.81&	83.12&	78.36&	48.26&	70.85\\
UniverSeg*    & \ding{51} & 73.25&	78.16&	84.57&	87.44&	82.23&	\underline{87.82}&	\underline{89.79}&	\underline{77.86}&	\textbf{88.57}&	\underline{90.28}&	\textbf{73.37}&	\underline{83.03}\\
Neuroverse3D & \ding{51} & \underline{85.69} & \textbf{83.98} & \textbf{89.98} & \underline{93.67} & \underline{82.66} & 75.75 & 78.08 & 74.66 & \underline{87.23} & 80.55 & 59.83 & 81.10 \\
\textbf{Medverse} & \ding{51} & \textbf{87.30} & \underline{82.12} & \underline{87.65} & \textbf{95.90} & \textbf{91.05} & \textbf{95.31} & \textbf{92.63} & \textbf{78.15} & 87.13 & \textbf{92.21} & \underline{70.48} & \textbf{87.27} \\
\bottomrule
\end{tabular}
\caption{Segmentation comparison across unseen centers, organs, species, and modalities in terms of Dice scores (\%). The context set and the training set for few-shot models are both of size 4. 2D models marked with an * utilize the orthogonal view fusion method.}
\label{tab:icl_comparison}
\end{table*}

\begin{table}[h]
\centering
\scriptsize
\setlength{\tabcolsep}{0.22pt}
\begin{tabular}{l|c|cc|cccc|c}
\toprule
\multirow{2}{*}{\textbf{Methods}}  & 
\multirow{2}{*}{\renewcommand{\arraystretch}{0.8}\begin{tabular}[x]{@{}c@{}}\textbf{Fine-}\\ \textbf{Tuning}\\\textbf{Free}\end{tabular}} & 
\multicolumn{2}{c|}{\textbf{\textit{Transformation}}} & \multicolumn{4}{c|}{\textbf{\textit{Enhancement}}} & \multirow{2}{*}{\textbf{Average}} \\
 & &
\renewcommand{\arraystretch}{0.8}\begin{tabular}[x]{@{}c@{}}Skull\\Stripping\end{tabular}  & 
\renewcommand{\arraystretch}{0.8}\begin{tabular}[x]{@{}c@{}}Modality\\Transform\end{tabular}  & 
\renewcommand{\arraystretch}{0.8}\begin{tabular}[x]{@{}c@{}}Bias\\ Removal\end{tabular}  & 
\renewcommand{\arraystretch}{0.8}\begin{tabular}[x]{@{}c@{}}Gaussian\\Noise\end{tabular}  & 
\renewcommand{\arraystretch}{0.8}\begin{tabular}[x]{@{}c@{}}Salt\\Pepper\end{tabular}  & 
\renewcommand{\arraystretch}{0.8}\begin{tabular}[x]{@{}c@{}} Inpain-\\ting\end{tabular} & \\
\midrule
\multicolumn{9}{l}{\textit{Fully Supervised Task-Specific Models (Upper Bound)}} \\
3D U-Net     & \ding{55} & 26.18&	32.97&	30.09&	31.86&	38.22	&34.57&	32.31\\
Swin-UNETR   & \ding{55} & 25.23 &	29.75 &	29.60 &	29.64 &	33.29 &	31.09 &	29.77\\
\midrule
\multicolumn{9}{l}{\textit{Few-Shot Task-Specific Models}} \\
3D U-Net     & \ding{55} & 22.78&	28.74	&27.34	&29.22&	34.65&	32.23&	29.16 \\
Swin-UNETR   & \ding{55} & 23.64 & 27.39 & 27.26 & 25.08 & 29.41 & 25.50 & 26.38 \\
\midrule
\multicolumn{9}{l}{\textit{ICL Models}} \\
Painter       & \ding{51} &  9.15 & 13.43 & 15.60 & 14.86 & 13.90 & 12.38 & 13.22 \\
Neuralizer    & \ding{51} & 21.01 & 24.71 & 26.13 & 13.14 & 25.26 & 23.75 & 22.34 \\
Painter*       & \ding{51} &  9.98&	12.98&	15.63&	14.74&	12.94&	14.20&	13.41 \\
Neuralizer*    & \ding{51} & 23.95&	\textbf{25.04}&	26.09&	19.10&	24.55&	25.28&	24.00 \\
Neuroverse3D  & \ding{51} & \underline{26.31} & 24.37 & \underline{28.42} & \underline{25.71} & \underline{30.65} & \underline{27.01} & \underline{27.08} \\
\textbf{Medverse} & \ding{51} &  \textbf{26.36} &  \textbf{24.80} &  \textbf{28.48} &  \textbf{26.08} &  \textbf{32.57} &  \textbf{31.51} &  \underline{28.30} \\
\bottomrule
\end{tabular}
\caption{Performance comparison across image transformation and enhancement tasks with a context size of 4 in terms of PSNR. Enhancement tasks are the average over 5 held-out datasets. 2D models marked with an * utilize the orthogonal view fusion method.}
\label{tab:perturbation_snp}
\end{table}

\section{Experiments}
\subsection{Data and Tasks}
\noindent\textbf{Datasets.}
To ensure robust cross-center generalization and data diversity, we curated a collection of \textbf{27 publicly available datasets} spanning multiple imaging modalities and acquisition centers, comprising a total of \textbf{40{,}362 3D scans}. The dataset encompasses widely used medical imaging modalities, including T1, T2, FLAIR, MRA, DWI, ADC, PD, and CT, as well as commonly studied anatomical regions such as the \textbf{brain}, \textbf{abdomen}, \textbf{prostate}, and \textbf{lung}. 22 datasets~\cite{yang2023benchmarking, CAS2023, hernandez2022isles, liew2022large, IXI, hoopes2022learning, marcus2007open, flanders2020construction, ADHD, jack2008alzheimer, alexander2017open, holmes2015brain, gera2023characterizing, nugent2022nimh, sudlow2015uk, menze2014multimodal, litjens2014evaluation, kuijf2019standardized, zeng2023imagecas, ji2022amos, luo2024rethinking, wasserthal2023totalsegmentator, antonelli2022medical} were used for training and validation with a random 9:1 split. The remaining five datasets were reserved as held-out sets to evaluate generalization to unseen distributions. These held-out datasets include brain and abdomen scans from previously unseen centers~\cite{marek2011parkinson, ma2024unleashing}, the nasal cavity as an unseen anatomical target~\cite{zhang2024nasalseg}, mice as an unseen species~\cite{rosenhain2018preclinical}, and PET as an unseen imaging modality~\cite{jack2008alzheimer}. Each held-out dataset was split in a 5:5 ratio into a meta context set used for semantic context selection and a test set.

\vspace{0.4em} \noindent\textbf{Data Preprocessing.} To improve the model's generalizability across acquisition centers, no spatial resampling was applied. Image intensities were normalized to the range from 0 to 1 using the 0.02 and 0.98 percentile values as lower and upper bounds, respectively. 
Segmentation masks were binarized by assigning zero to the background and one to the foreground.

\vspace{0.4em}
\noindent\textbf{Tasks.}  
We trained our model on three categories of tasks, encompassing segmentation, transformation, and enhancement objectives. The segmentation tasks include anatomical structure delineation across various organs, as well as tumor and vessel segmentation. The transformation tasks include arbitrary modality-to-modality transformation and brain extraction for skull stripping. The enhancement tasks involve improving image quality through bias field correction, image inpainting, and the removal of Gaussian and salt-and-pepper noise.

Further details regarding datasets, task definitions, training procedures, and evaluation protocols are provided in the supplemental material.

\begin{figure*}
\centering
\includegraphics[width=0.95\textwidth]{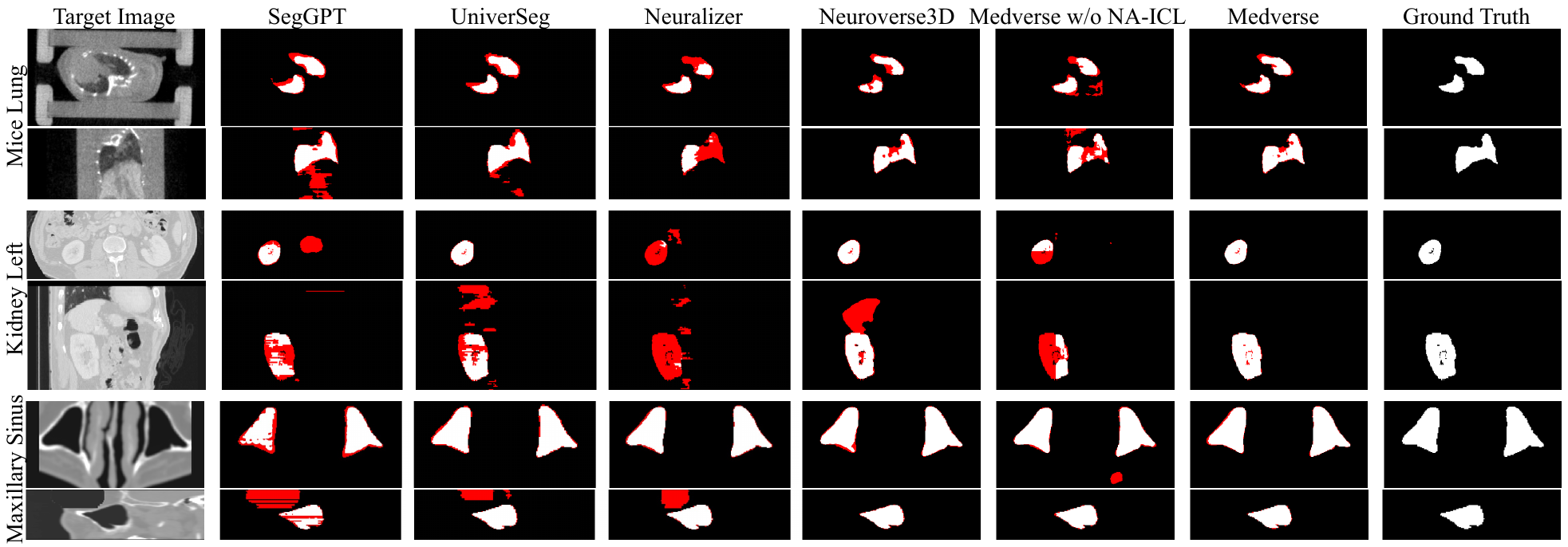}
\caption{Qualitative results of ICL models on segmentation tasks. For each 3D segmentation target, results are shown from two views. Red regions indicate segmentation errors. The 2D models take the slice corresponding to the first view as input. Medverse w/o NA-ICL denotes a variant of Medverse without autoregressive processing.}
\label{fig:qualitative1}
\end{figure*}

\subsection{Compared Models}

\noindent\textbf{Medverse.}  
Both the target and context branches are implemented using a five-stage 3D U-Net architecture~\cite{cciccek20163d}.  
Each branch begins with 32 channels in the first stage, with the number of channels doubling at each subsequent stage. The model operates on input patches of size $128 \times 128 \times 128$. In BAM, $p$ is set to 4, and $m$ is set to 66.

\vspace{0.4em}
\noindent\textbf{Task-Specific Models.}  
We compared Medverse against task-specific 3D models.  
Each task-specific model was trained directly on the meta context set of the held-out datasets, thereby mitigating domain shift concerns.  
This setting reflects the conventional practice in clinical environments, where models are trained specifically for a given task.  
Two backbone architectures were evaluated: a 3D U-Net, which shares the same architecture and channel configuration as the Medverse backbone, and the Swin-UNETR~\cite{hatamizadeh2021swin}.  
We assessed performance under both few-shot and fully supervised learning scenarios.

\vspace{0.4em}
\noindent\textbf{Other ICL Models.}  
We compared our method with several state-of-the-art ICL approaches, including Painter~\cite{wang2023images}, SegGPT~\cite{wang2023seggpt}, UniverSeg~\cite{butoi2023universeg}, and Neuralizer~\cite{czolbe2023neuralizer}, all of which are designed for 2D inputs.  
To adapt to 3D, we split each 3D target image into 2D slices and randomly sampled slices from the 3D context set containing the target region to construct the 2D context set.  
The number of context slices was set to 1 for Painter, 8 for SegGPT, 32 for Neuralizer, and 64 for UniverSeg, following the optimal configurations reported in their respective publications.  
The resulting 2D outputs were reassembled into 3D volumes for metric evaluation.  
2D models marked with an asterisk (*) utilize the orthogonal view fusion method from 3 views, which introduces a three-fold increase in computational load.
We also compared with Neuroverse3D~\cite{hu2025building}, an ICL model developed for 3D neuroimaging.  
Due to its limited input resolution of $128 \times 128 \times 128$, we resized the input images to fit Neuroverse3D and rescaled the outputs back to the original resolution for evaluation.  
All models were evaluated using their publicly released pretrained weights.

\begin{figure}[h]
\centering
\includegraphics[width=0.45\textwidth]{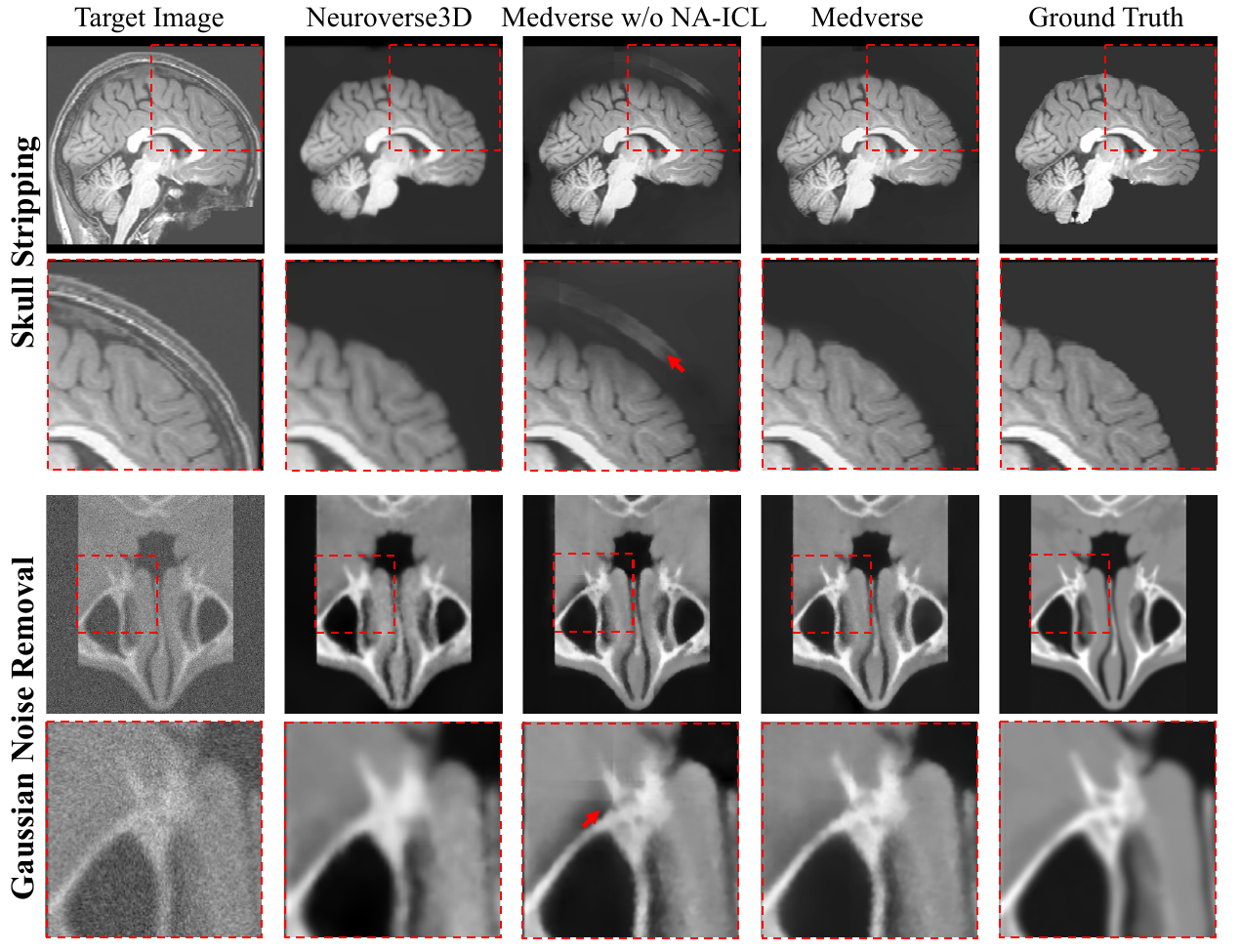}
\caption{Qualitative results of 3D ICL models. Medverse w/o NA-ICL denotes the variant of Medverse without autoregressive context. The second row for each task presents zoomed-in views highlighting differences in generated details. Neuroverse3D produces outputs with limited resolution. Red arrows indicate artifacts.}
\label{fig:qualitative2}
\end{figure}

\subsection{Results on Held-Out Dataset}

\vspace{0.4em}
\noindent\textbf{Quantitative Comparison on Segmentation Tasks.}  
Table~\ref{tab:icl_comparison} reports the segmentation performance of the models. Medverse achieves the highest performance among all ICL methods, surpassing the second-best method, Neuroverse3D, by about 6 points in the average Dice score. This improvement can be attributed to Medverse's broader training data coverage and its superior ability to capture fine details enabled by the NA-ICL framework. 2D models, such as UniverSeg and SegGPT, exhibit decreased performance on 3D volumes, primarily due to their tendency to produce false positives in slices that do not contain the target anatomy. When compared to few-shot task-specific models, Medverse achieves similar levels of performance while offering the notable advantage of requiring no task-specific fine-tuning. These results highlight the strong potential and practical value of our method. We further evaluate our model across more classes on the validation set, with results provided in the Supplementary Table 4.

\vspace{0.4em}
\noindent\textbf{Quantitative Comparison on Transformation and Enhancement Tasks.}  
Table~\ref{tab:perturbation_snp} presents the performance of different models on transformation and enhancement tasks. Medverse significantly outperforms other ICL models on tasks such as inpainting and denoising. Although Medverse achieves only a modest improvement of 1.22 PSNR over the second-best Neuroverse3D on average, it preserves the full resolution of the input image, whereas Neuroverse3D limits the resolution of its outputs. Despite being the best among ICL models, Medverse still lags behind the fully supervised 3D U-Net and shows a performance gap compared to the few-shot 3D U-Net.

\vspace{0.4em}
\noindent\textbf{Qualitative Results on Segmentation Tasks.}  
Figure~\ref{fig:qualitative1} presents the segmentation results of different ICL models. The 2D models perform reasonably well on individual slices that contain the target structure but exhibit discontinuities across adjacent slices, and SegGPT and UniverSeg tend to produce false positives in slices without the target. Neuroverse3D alleviates the limitations of 2D processing but struggles to generalize to unseen organ. Medverse w/o NA-ICL refers to applying a sliding window directly, without autoregressive context. In this case, the model lacks access to sufficient global context, leading to unstable predictions. In contrast, Medverse yields the most accurate and fine-grained segmentation results.

\vspace{0.4em}
\noindent\textbf{Qualitative Results on Transformation and Enhancement Tasks.}  
Figure~\ref{fig:qualitative2} illustrates the performance of 3D ICL models on skull stripping and Gaussian noise removal tasks. As shown, Neuroverse3D fails to preserve image resolution, resulting in a loss of fidelity, whereas Medverse effectively retains detailed structures. Without the NA-ICL framework, the model struggles to accurately remove the skull, as having access only to local information increases task difficulty. In the Gaussian noise removal task, stitching artifacts appear as visible lines, which are substantially mitigated when the NA-ICL framework is employed.

Additional qualitative results on segmentation, transformation, and enhancement can be found in the supplementary material.

\begin{figure}
\centering
\includegraphics[width=0.473\textwidth]{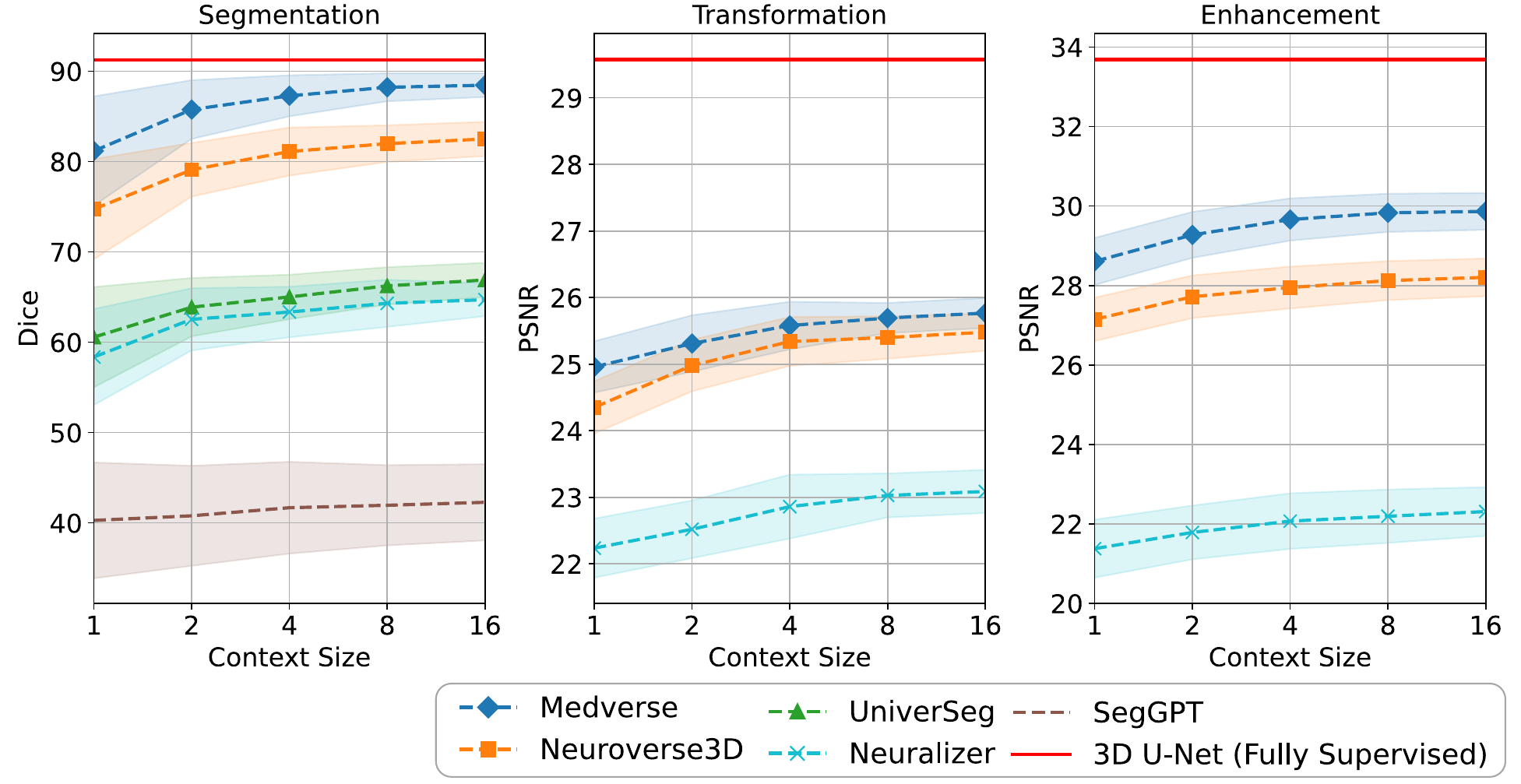}
\caption{Performance comparison of ICL models under varying 3D context sizes.}
\label{fig:context_size}
\end{figure}

\vspace{0.4em}
\noindent\textbf{Effect of Context Size.}  
Figure~\ref{fig:context_size} shows how the performance of different ICL models varies with context size. As expected and consistent with prior findings, the performance of all models improves as the context size increases. For Medverse, the Dice score on segmentation tasks improves by 7.28 points with an increasing context size from 1 to 16, while PSNR increases by 0.81 and 1.26 for transformation and enhancement tasks, respectively. The more substantial improvement on segmentation tasks may be attributed to the sparsity of segmentation labels. In contrast, transformation and enhancement tasks typically provide richer semantic guidance per sample, so the marginal benefit from increasing context size is smaller. Furthermore, Medverse consistently outperforms other ICL models across all three kinds of tasks. In transformation tasks, Neuroverse3D performs comparably to Medverse since both skull stripping and modality transformation use neuroimaging data within its training domain. Its slightly lower performance mainly stems from reduced output resolution, unlike Medverse, which preserves full resolution.

\vspace{0.4em}
\noindent\textbf{Computational Efficiency.}  
BAM processes only $4^3$ tokens, adding just $2.01\times10^{-1}$ GFLOPs and $9.18\times10^{-4}$ memory overhead per context pair. In contrast, standard cross-attention with $128^3$ tokens incurs $2.36\times10^6$ GFLOPs and $2.55$ memory overhead, making it computationally impractical. Additionally, our model maintains a fixed inference memory usage of 9.14 GB regardless of context size, enabled by adaptive parallel-sequential context processing. Inference on a $128\times128\times128$ patch with 8 context samples and 1 autoregressive context takes only 1.16 seconds, whereas models like Painter and SegGPT exceed 30 seconds. These properties make our model both efficient and practical for deployment. Full details are provided in the supplemental material.

\vspace{0.4em}
\noindent\textbf{Ablation Analysis of Key Components.}  
Table~\ref{tab:ablation_bam_autoreg} shows ablations on the BAM module and the NA-ICL framework. Removing NA-ICL, which replaces coarse-to-fine prediction with direct sliding-window inference, leads to a notable drop in segmentation performance. This is due to the limited perspective field of the target image and the loss of global context guidance. For transformation and enhancement tasks, NA-ICL also improves performance, with one contributing factor being its ability to enhance consistency across sliding-window patches. When the BAM module is removed, feature fusion is instead performed via simple concatenation as in~\cite{hu2025building}. The BAM module leads to overall performance improvements due to mitigating the adverse effects caused by spatial misalignment between the context and target images.

\begin{table}[htbp]
\centering
\scriptsize
\setlength{\tabcolsep}{6pt}
\begin{tabular}{cc|ccc}
\toprule
\textbf{BAM} & \textbf{NA-ICL} & \textbf{Seg. (Dice ↑)} & \textbf{Tran. (PSNR ↑)} & \textbf{Enh. (PSNR ↑)} \\
\midrule
             &                & 76.50 & 24.53 & 27.45 \\
\ding{51}    &                & 78.90 & 25.38 & 27.93 \\
\ding{51}    & \ding{51}      & \textbf{87.27} & \textbf{25.58} & \textbf{29.66}   \\
\bottomrule
\end{tabular}
\caption{Ablation study on the BAM module and NA-ICL framework. BAM is replaced with plain feature concatenation when removed, and NA-ICL is replaced with sliding window processing.}
\label{tab:ablation_bam_autoreg}
\end{table}

\section{Conclusion}
In this work, we introduce Medverse, a universal in-context learning model for 3D medical image analysis that performs segmentation, transformation, and enhancement tasks across diverse organs, modalities, and centers. Trained on large-scale heterogeneous data, Medverse generalizes well to unseen domains without task-specific fine-tuning. To enable high-fidelity predictions, we propose the next-scale autoregressive framework for multi-scale anatomical awareness and full-resolution outputs, along with a blockwise Cross-Attention Module for efficient long-range context–target interaction. Extensive evaluations show that Medverse not only outperforms existing ICL models but also establishes a new paradigm for universal 3D medical image processing, effectively addressing the challenge of high-resolution data and revealing the broader potential of ICL in medical imaging.

\noindent\textbf{Limitation.}  Due to limited resources, the current model uses a relatively modest number of parameters and a constrained number of training datasets, leaving room for future scaling to achieve even better performance. Furthermore, conflicts between different tasks may not have been fully resolved. For instance, exclusively utilizing the segmentation Dice loss while discarding the transformation and enhancement tasks could potentially yield superior segmentation results. Therefore, how to more effectively unify the transformation and enhancement tasks with the segmentation task warrants further investigation.


\bibliography{aaai2026}

\clearpage
\onecolumn


\begin{figure*}[t]
\centering
\Large 
\textbf{Medverse: A Universal Model for Full-Resolution 3D Medical Image Segmentation, Transformation and Enhancement} \\[0.5em]
Supplementary Material \\[1.0em]
\end{figure*}

\section{Related Work}
\label{sec:related}

\subsection{Domain Diversity in Medical Imaging}

Medical imaging data exhibits high variability across multiple axes, including anatomical regions, imaging modalities (e.g., CT, MRI, PET, Ultrasound), acquisition settings, and institutions~\cite{bell2022harmonization, zhang2020generalizing}. This diversity poses significant challenges for conventional models, which are typically trained on narrow, task-specific datasets, resulting in poor generalization to unseen domains. Researchers have explored domain adaptation approaches~\cite{wang2023fvp, hu2024chebyshev}, which involve fine-tuning models on the target domain. However, this requirement limits their scalability and makes wide deployment in clinical settings difficult. Domain generalization methods have also been proposed to overcome this limitation without requiring target-domain fine-tuning. These methods either expand the training distribution via predesigned strategies~\cite{zhang2020generalizing, ouyang2022causality}, or attempt to learn domain-invariant features~\cite{hu2022domain, ilse2020diva}. Nevertheless, their effectiveness in overcoming domain shift remains limited and their performance gains are often unstable. Recently, training universal models on large-scale and diverse datasets has emerged as a promising solution to address domain shift in medical imaging~\cite{butoi2023universeg, ma2024segment, hu2025building}. These models demonstrate strong generalization capabilities across institutions, imaging modalities, and even tasks, making them ideal choices for deployment in clinical centers.

\subsection{In-Context Learning in Medical Imaging}

In-Context Learning (ICL) has emerged as a powerful paradigm for performing multiple tasks by conditioning inference on a small set of examples, known as contexts, without modifying model parameters. ICL was first introduced in the field of natural language processing~\cite{brown2020language}, and has been shown to generalize effectively to out-of-distribution data~\cite{reddy2023mechanistic, chan2022data}. As a result, ICL is increasingly viewed as a promising approach for overcoming domain shift in computer vision. Models such as SegGPT~\cite{wang2023seggpt} and Painter~\cite{wang2023images} have demonstrated the potential of ICL across a wide range of vision tasks. In the medical imaging domain, UniverSeg and Neuralizer—based on convolutional and multi-branch architectures—have validated the effectiveness of ICL for 2D segmentation tasks and 2D neuroimaging tasks, respectively. ICL-SAM~\cite{hu2024icl} combines the ICL model with SAM, further improving the performance of ICL models on 2D medical imaging segmentation tasks. Building on these works, Neuroverse3D further extended ICL to the 3D setting while reducing the memory burden of ICL modules, demonstrating promising results in 3D neuroimaging applications. However, Neuroverse3D remains limited to neuroimaging and struggles to handle high-resolution 3D medical images, which constrains its applicability in broader clinical scenarios. Building ICL frameworks that can apply to diverse medical scenarios and high-resolution 3D images remains an open challenge.

\subsection{Universal Models in Medical Imaging}
A growing number of universal models have been developed in medical imaging, particularly for segmentation tasks~\cite{ma2024segment, wong2024scribbleprompt,liu2023clip, ye2023cardiacseg}. Prompt-based universal models include those that utilize point and bounding box inputs, such as SAM~\cite{kirillov2023segment} and MedSAM~\cite{ma2024segment}, as well as ScribblePrompt~\cite{wong2024scribbleprompt}, which incorporates scribble annotations, and nnInteractive, which extends these methods to 3D. These approaches have shown promising results. However, a key difference is that such models require manual input for each image, such as interactive annotations, whereas ICL models require no manual intervention once context examples are provided. These two types of models can be applied in different scenarios—or even combined, as explored in~\cite{hu2024icl}. Moreover, since ICL models take image-label pairs as input, they can be naturally extended to support a wider range of tasks, including image transformation and enhancement. Another line of work has explored natural language prompts to instruct models, such as CLIP-based frameworks adapted to the medical domain~\cite{liu2023clip, liu2024universal}. While these models exhibit strong generalization capabilities, their reliance on textual input makes it difficult for them to generalize beyond pre-defined targets or tasks seen during training.

\clearpage
\section{Autoregressive Inference Pipeline of Medverse}
\begin{algorithm}[h]
\caption{Autoregressive Inference Pipeline of Medverse}
\label{alg:medverse_inference}
\begin{algorithmic}[1]  
\Require Target volume $\bm{x}\in\mathbb{R}^{H\times W\times D}$; 
        semantic context $S\in\mathbb{R}^{L\times 2 \times H\times W\times D}$ 
        (where $L$ is the context size and the second dimension indexes the image–label pairs); 
        patch size $I$; 
        prediction network $F(\cdot)$
\Ensure  Final prediction $\hat{\bm{y}}\in\mathbb{R}^{H\times W\times D}$
\State $T \gets \bigl\lceil\log_{2}\bigl(\frac{\max\{H,W,D\}}{I}\bigr)\bigr\rceil + 1$
       \Comment{number of autoregressive steps}
\State $\mathcal{A}^{(0)} \gets \varnothing$ \Comment{empty autoregressive context}
\For{$t = 1$ \textbf{to} $T$}                       \Comment{coarse $\rightarrow$ fine}
    \State $H^{(t)}\!\gets\!\lceil H / 2^{\,T-t}\rceil,\;
           W^{(t)}\!\gets\!\lceil W / 2^{\,T-t}\rceil,\;
           D^{(t)}\!\gets\!\lceil D / 2^{\,T-t}\rceil$
    \State Resize $\bm{x}$ and $S$ to $\bigl(H^{(t)},W^{(t)},D^{(t)}\bigr)$
    \If{$t == 1$}          \Comment{global low-resolution prediction}
        \State $\hat{\bm{y}}^{(1)} \gets
               F\bigl(\bm{x}^{(1)},\, S^{(1)},\, \mathcal{A}^{(0)}\bigr)$
    \Else                   \Comment{patch-wise refinement}
        \State Divide $\bm{x}^{(t)}$ into (possibly overlapping) patches
               $\{\bm{x}^{(t)}_i\}$ of size $I^3$
        \ForAll{patch index $i$}
            \State Crop aligned semantic context $S^{(t)}_i$
            \State Extract corresponding region
                   $\mathcal{A}^{(t-1)}_i$ from $\mathcal{A}^{(t-1)}$
                   and upsample to $I^3$
            \State $\hat{\bm{y}}^{(t)}_i \gets
                   F\bigl(\bm{x}^{(t)}_i,\,
                          S^{(t)}_i,\,
                          \mathcal{A}^{(t-1)}_i\bigr)$
        \EndFor
        \State Aggregate $\{\hat{\bm{y}}^{(t)}_i\}$
               to obtain $\hat{\bm{y}}^{(t)}$
    \EndIf
    \State $\mathcal{A}^{(t)} \gets \bigl(\bm{x}^{(t)},\, \hat{\bm{y}}^{(t)}\bigr)$
           \Comment{update AR context}
\EndFor
\Return $\hat{\bm{y}}^{(T)}$
\end{algorithmic}
\end{algorithm}

\section{Experiments Setting}
\label{sec:data_sup}

\begin{table*}[h!]
    \centering
    \renewcommand{\arraystretch}{1.1}
    \setlength{\tabcolsep}{4pt}
    \resizebox{0.85\textwidth}{!}{
    \begin{tabular}{cccccc>{\centering\arraybackslash}p{3cm}}
        \toprule
        \textbf{Type for use} & \textbf{Dataset}& \textbf{Organ} & \textbf{Task} & \textbf{\# Scans} & \textbf{\# Masks} & \textbf{Modality} \\
        \midrule
        \multirow{17}{*}{\begin{tabular}[x]{@{}c@{}}Training and \\Validation Set\end{tabular}} 
        & TopCow~\cite{yang2023benchmarking} & Brain & Seg., Enh. & 90 & 90 & MRA \\
        & CAS2023~\cite{CAS2023} & Brain & Seg., Enh.  & 100 & 100 & MRA \\
        & ISLES2022~\cite{hernandez2022isles} & Brain & Enh., Tran.  & 750 & 0 & \renewcommand{\arraystretch}{0.8}\begin{tabular}[x]{@{}c@{}}DWI, ADC,\\FLAIR\end{tabular}\\
        & ATLAS~\cite{liew2022large} & Brain & Seg., Enh. & 655 & 655 & T1w \\
        & IXI~\cite{IXI} & Brain & Enh., Tran.  & 2268 & 0 &   \renewcommand{\arraystretch}{0.8}\begin{tabular}[x]{@{}c@{}}T1, T2,\\MRA, PD\end{tabular}\\
        & ICH Unlabeled~\cite{flanders2020construction} & Brain & Enh.  & 2000 & 0 & CT \\
        & ADHD~\cite{ADHD} & Brain & Seg., Enh., Tran.  & 950 & 200 & T1 \\
        & ADNI~\cite{jack2008alzheimer} & Brain & Seg., Enh., Tran. & 9923 & 200 & T1 \\
        & CMI~\cite{alexander2017open} & Brain & Seg., Enh., Tran.  & 5146 & 200 & T1 \\
        & GSP~\cite{holmes2015brain} & Brain & Seg., Enh., Tran.  & 2616 & 200 & T1 \\
        & HAB~\cite{gera2023characterizing} & Brain & Seg., Enh., Tran.  & 460 & 460 & T1 \\
        & NIMH~\cite{nugent2022nimh} & Brain & Seg., Enh., Tran.  & 248 & 248 & T1 \\
        & OASIS~\cite{marcus2007open} & Brain & Seg., Enh., Tran.  & 3916 & 828 & T1 \\
        & UKBiobank~\cite{sudlow2015uk} & Brain & Seg., Enh.  & 4000 & 2000 & T1, T2 \\
        & BraTS~\cite{menze2014multimodal} & Brain & Seg., Enh., Tran.  & 5004 & 1251 & \renewcommand{\arraystretch}{0.8}\begin{tabular}[x]{@{}c@{}}FLAIR, T1,\\T1CE, T2\end{tabular} \\
        & WMH~\cite{kuijf2019standardized} & Brain & Enh., Tran. & 120 & 0 & T1, FLAIR \\
        & PROMISE12~\cite{litjens2014evaluation} & Prostate & Seg., Enh. & 50 & 50 & T2 \\
        & CAS~\cite{zeng2023imagecas} & Cardiac & Seg., Enh. & 190 & 190 & CTA \\
        & AMOS22~\cite{ji2022amos} & Abdomen & Seg., Enh. & 360 & 360 & CT \\
        & RAOS~\cite{luo2024rethinking} & Abdomen & Seg., Enh. & 317 & 317 & CT \\
        & TotalSeg~\cite{wasserthal2023totalsegmentator} & Abdomen & Seg., Enh. & 87 & 87 & CT \\
        & MSD~\cite{antonelli2022medical} & Lung, Cardiac & Seg., Enh. & 83 & 83 & CT \\

        \midrule
        \multirow{5}{*}{Held-out Set} 
        & PPMI~\cite{marek2011parkinson} & Brain & Seg., Enh., Tran.  & 220 & 220 & T1 \\
        & FLARE22~\cite{ma2024unleashing} & Abdomen & Seg., Enh.  & 50 & 50 & CT \\
        & Nasal~\cite{zhang2024nasalseg} & Nasal & Seg., Enh.  & 130 & 130 & CT \\
        & Mice~\cite{rosenhain2018preclinical} & Body & Seg., Enh.  & 40 & 40 & CT \\
        & ADNI\cite{jack2008alzheimer} & Brain & Seg., Enh.  & 589 & 589 & PET \\
        \midrule

        & \textbf{Total} & \renewcommand{\arraystretch}{0.8}\begin{tabular}[x]{@{}c@{}}Brain, Lung,\\ Abdomen, Cardiac\\ Prostate, Nasal\end{tabular} & Seg., Enh., Tran. & 40362 & 8548 & \renewcommand{\arraystretch}{0.8}\begin{tabular}[x]{@{}c@{}}T1, T2, FLAIR,\\MRA, DWI, ADC,\\ PD, CT, PET\end{tabular} \\
        
        \bottomrule
    \end{tabular}}
    \caption{Summary of datasets. Segmentation, enhancement, and transformation tasks are denoted as Seg., Enh., and Tran., respectively. The PET modality data from the ADNI dataset was not included in training.}

    \label{tab:dataset_summary}
\end{table*}

\begin{figure}[h]
\centering
\includegraphics[width=0.45\textwidth]{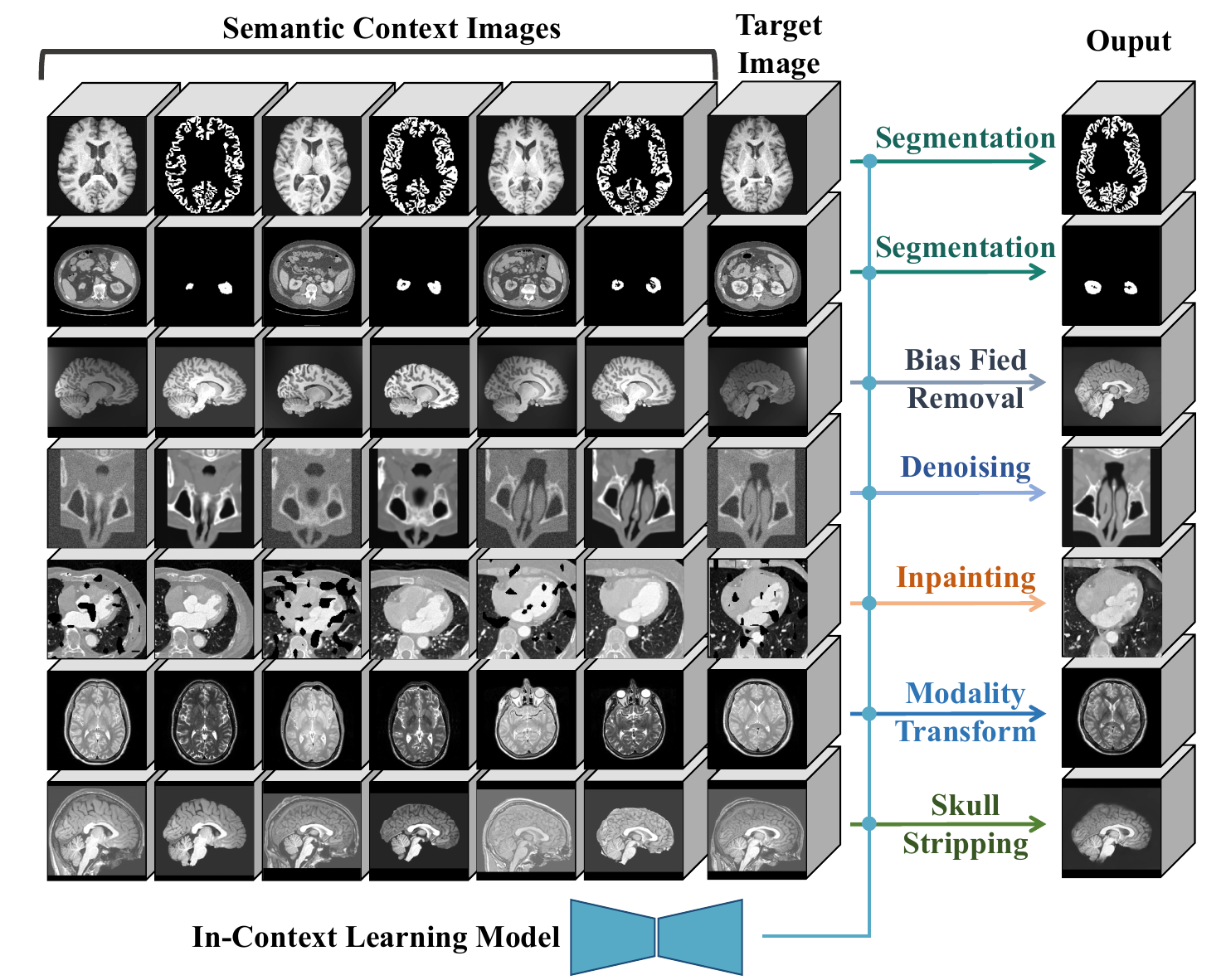}
\caption{An overview of tasks executable by the model.}
\label{fig:intro}
\end{figure}

\subsection{Data Sampling}
\label{sec:data_sup_}

During training, tasks are selected according to a predefined sampling distribution.  
For each selected task, a dataset is randomly chosen with equal probability.  
Given a context size of $L$, we randomly sample $L+1$ image–label pairs from the training split of the selected dataset.  
One of these samples is alternately assigned as the target image and corresponding ground truth, while the remaining $L$ samples constitute the context set, following the same protocol as described in~\cite{hu2025building}.  
The sampling distribution over tasks is summarized in Table~\ref{tab:task_sampled_weight}.

\begin{table}
\centering
\renewcommand{\arraystretch}{1} 
\resizebox{0.47\textwidth}{!}{
    \begin{tabular}{ccc}
    \toprule
    \textbf{Task} & \textbf{Sampled Rate} & \textbf{Weight} \\
    \midrule
    Segmentation & 3 & 50 \\
    Bias Remove & 2 & 1 \\
    Gaussian Noise Remove & 2 & 1 \\
    Salt \& Pepper Noise Remove & 2 & 1 \\
    Inpainting & 2 & 1 \\
    Skull Stripping & 1 & 1 \\
    Modality Transform & 1 & 1 \\
    \bottomrule
    \end{tabular}
    }
    \caption{Sampling rate and weight for each task during training.}
\label{tab:task_sampled_weight}
\end{table}

\subsection{Task Construction}
\label{sec:tasks_sup}

\vspace{0.4em}\noindent\textbf{Segmentation:} Our model performs binary segmentation. For multi-class segmentation datasets, each class is treated as the foreground in turn, while all other classes are merged into the background.  For prediction, binary masks were produced by thresholding the model outputs at 0.5. 

\vspace{0.4em}\noindent\textbf{Skull Stripping:} The objective is to remove the skull region from brain images~\cite{hoopes2022synthstrip}. Ground-truth skull-stripping masks are generated using FreeSurfer~\cite{fischl2012freesurfer} from T1-weighted brain images. For this task, we performed evaluation on a held-out subset of the PPMI dataset.

\vspace{0.4em}\noindent\textbf{Modality Transformation:} This task converts images from one modality to another, where the input and output volumes are spatially registered. For this task, we evaluated on the PD$\rightarrow$T2 modality transformation task from the IXI dataset, as the modalities in this dataset are well spatially aligned, making it suitable for evaluation. Moreover, the PD$\rightarrow$T2 task was excluded from the training tasks.

\vspace{0.4em}\noindent\textbf{Bias-Field Correction:} The goal is to correct intensity inhomogeneities in MRI scans~\cite{goldfryd2021deep}. Three-dimensional bias fields are synthesized using Legendre polynomials with randomly sampled coefficients. For all enhancement tasks, we perform evaluation on five held-out sets and report the average results.

\vspace{0.4em}\noindent\textbf{Gaussian Noise Removal:} This task is synthesized by adding zero-mean Gaussian noise with a standard deviation uniformly sampled from the range 0.15 to 0.25.

\vspace{0.4em}\noindent\textbf{Salt-and-Pepper Noise Removal:} This task is synthesized by introducing salt noise (intensity 1) and pepper noise (intensity 0) independently with a probability of 0.04.

\vspace{0.4em}\noindent\textbf{Inpainting:} The objective is to reconstruct missing or occluded regions in the input image~\cite{liu2021symmetric}. This task is synthesized using binary masks generated from random three-dimensional Perlin noise to occlude regions of the input.

\subsection{Image and Task Augmentation}
\label{sec:aug}

Image and task augmentation are critical for training in-context learning models and are applied after data sampling.  
For image augmentation, we applied the following transformations: random affine transformations (probability 0.05), elastic deformations (0.05), flipping (0.05), and rotations (0.05).  
To further increase image diversity, we introduced random intensity shifts (0.2), intensity scaling (0.2), Gaussian noise (0.1), and intensity inversion (0.05).

For task augmentation, we employed several strategies, including Sobel filtering, mask inversion, random dilation, random erosion, task overlapping, and foreground randomization, following the procedures described in~\cite{hu2025building}.

\vspace{0.4em}
\noindent\textbf{GIN.}
We also applied a random convolutional network known as GIN~\cite{ouyang2022causality} (0.05), which generates images with randomized contrast.

\vspace{0.4em}
\noindent\textbf{Synthetic Data.}
We incorporated synthetic data following~\cite{butoi2023universeg, hoffmann2021synthmorph}.  
We used randomly sampled binary masks, brain masks, and abdomen masks derived from TotalSegmentator~\cite{wasserthal2023totalsegmentator} to generate synthetic images.  
This process resulted in 150 synthetic segmentation datasets with varying contrast properties, each containing 100 three-dimensional image samples.

\subsection{Training}
During training, we adopt a teacher forcing approach, where the autoregressive context is replaced with ground-truth image-label pairs downsampled by a factor of 2. To prevent the model from over-relying on the autoregressive context—since it may be imperfect during actual inference—we introduce noise by using a larger downsampling factor. Specifically, instead of consistently using a factor of 2, we randomly sample the downsampling factor from the set \{2, 3, 4\} for the ground truth.

Medverse was trained on 2 NVIDIA A100 80GB GPUs with a batch size of 1 per GPU, using the ADAM optimizer.  
Training was conducted for 240K steps with an initial learning rate of $3\times 10^{-5}$.  
Validation loss was evaluated every 2.4K steps, and the learning rate was reduced by half if no improvement was observed over 20 consecutive evaluations.  
To improve training efficiency, the context size and mini-context size were fixed at three for the first 200K steps, requiring the model to process only a single mini-context.  
For the remaining 40K steps, the context size was uniformly sampled between 1 and 8.  
Training was completed in approximately 8 days, and the model with the lowest validation loss was selected for evaluation.
All task-specific models were trained on a single GPU for 36{,}000 steps due to the significantly smaller training sets and faster convergence.

\subsection{Evaluation}
We evaluate model performance using two standard metrics in medical image analysis: Dice coefficient and Peak Signal-to-Noise Ratio (PSNR). Dice is used for segmentation tasks and measures the spatial overlap between the predicted and ground truth masks. It is particularly suitable for medical segmentation due to its robustness to class imbalance. PSNR is used for image transformation and enhancement tasks. It quantifies the pixel-level similarity between the predicted and reference images, reflecting the fidelity of the output. Together, these metrics provide a comprehensive evaluation of both structural accuracy and image quality. Each task and setting was repeated 8 times with randomly sampled context sets.  
The average and standard deviation across runs were reported to ensure robust performance estimation.

\section{Additional Experiments}

\begin{figure}
\centering
\includegraphics[width=0.7\textwidth]{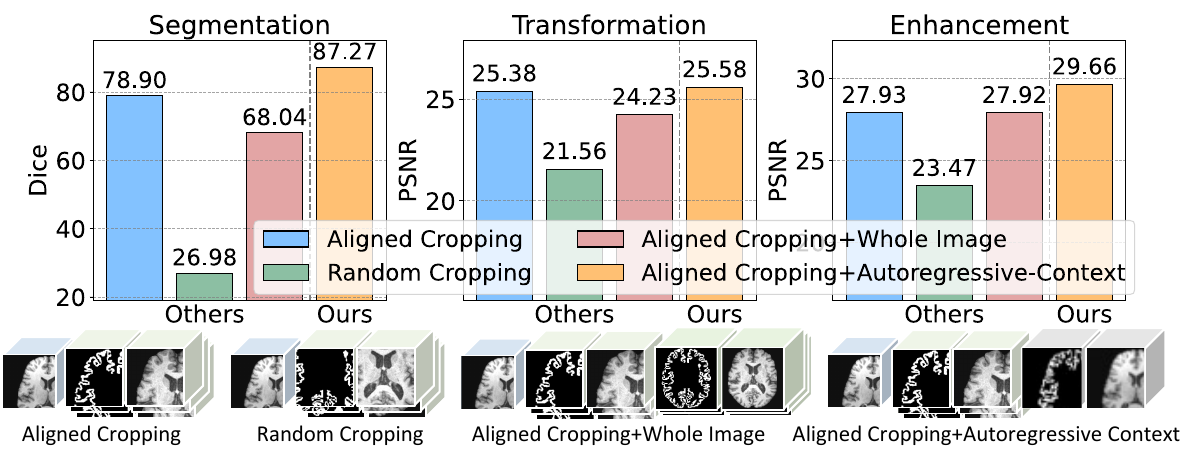}
\caption{Comparison of different context construction strategies during sliding-window inference. Aligned Cropping+Autoregressive Context represents our proposed method. Random Cropping selects crops randomly from foreground regions. Whole Image adds downsampled full images as context.}
\label{fig:context_con}
\end{figure}

\vspace{0.4em}
\noindent\textbf{Comparison of Context Construction Strategies.}  
Figure~\ref{fig:context_con} shows the impact of various context construction strategies. NA-ICL introduces an additional autoregressive context, forming a novel strategy. We fine-tuned models under corresponding strategies to ensure a fair comparison. Compared to aligned cropping, random cropping significantly reduces performance, indicating that anatomical alignment across samples is crucial for effective context learning. Directly incorporating the whole image as global context does not improve performance and can even degrade it suggesting that it is difficult to leverage coarse-resolution guidance directly across scales. Incorporating autoregressive context consistently improves performance across all kinds of tasks, highlighting the strength of our approach.

\vspace{0.4em}
\noindent\textbf{Comparison of the computational resources of BAM.}
Table~\ref{tab:compute} compares the computational cost of the BAM module and standard cross-attention. As shown in the table, BAM achieves a substantial reduction in computation, particularly at Stage 1. This is because the spatial size of the features in Stage 1 is large, leading to millions of tokens if standard cross-attention is used, which results in an extremely high computational cost. In contrast, BAM maintains a fixed number of tokens at \( p^3 \), regardless of spatial size. Overall, BAM consistently maintains lower FLOPs and memory usage, while still enabling long-range interactions. At the bottom of the table, we observe that increasing \( p \) gradually improves performance, reaching an optimal value around \( p = 4 \). This suggests two key insights. First, long-range interactions are essential for providing broader receptive fields, which are particularly beneficial for segmentation tasks. Second, in the medical domain, moderate levels of long-range attention are sufficient, and excessively dense attention may not yield additional gains.

\begin{table*}[htbp]
\centering
\caption{Computation cost, FP32 inference time on A100, and memory consumption during inference of the BAM module across different 3D U-Net stages. The cross-attention refers to standard attention without spatial sparsity. When BAM is configured with \( p = 1 \), the module is equivalent to direct concatenation and thus introduces no additional computation. The bottom of the table compares model performance with different values of \( p \). Due to the prohibitively high computational cost of standard cross-attention, its performance could not be reported. "Total" refers to the combined cost of fusion modules in both the encoder and decoder for 1 context pair.}
\label{tab:compute}
\setlength{\tabcolsep}{6pt}
\begin{tabular}{l l r r r r r}
\toprule
\begin{tabular}[c]{@{}l@{}}Stage\\  (Feature Size) \end{tabular} &
Metric & 
\begin{tabular}[c]{@{}l@{}}BAM\\  $p=1$ \end{tabular}    &
\begin{tabular}[c]{@{}l@{}}BAM\\  $p=2$ \end{tabular}    &
\begin{tabular}[c]{@{}l@{}}BAM\\  $p=4$ \end{tabular}    &
\begin{tabular}[c]{@{}l@{}}BAM\\  $p=8$ \end{tabular}    &
Cross-Attention \\
\midrule
\multirow{4}{*}{\begin{tabular}[c]{@{}l@{}}Stage 1\\ 32$\times$128$\times$128$\times$128\end{tabular}}
& GFLOPs & --	& $6.72\times10^{-2}$&$6.85\times10^{-2}$& $1.39\times10^{-1}$& $1.16\times10^{6}$ \\
& Time (s)  &--	& $3.44\times10^{-6}$ & $3.51\times10^{-6}$ & $7.10\times10^{-6}$ & $5.95\times10^{1}$\\
& Memory (GB) & --	& $1.16\times10^{-5}$  & $3.38\times10^{-5}$ & $2.70\times10^{-4}$ & 1.10 \\
& \# Token & $1$ & $2^3$ & $4^3$ & $8^3$ & $128^3$ \\
\midrule
\multirow{4}{*}{\begin{tabular}[c]{@{}l@{}}Stage 2\\ 64$\times$64$\times$64$\times$64\end{tabular}}
& GFLOPs & --	& $1.69\times10^{-2}$ & $1.84\times10^{-2}$ & $9.03\times10^{-2}$ & $1.81\times10^{4}$ \\
& Time (s)   & --	& $8.65\times10^{-7}$ & $9.44\times10^{-7}$  & $4.63\times10^{-6}$ & $9.3\times10^{-1}$\\
& Memory (GB) & --	& $2.11\times10^{-5}$ & $5.02\times10^{-5}$ & $2.83\times10^{-4}$ & $1.4\times10^{-1}$\\
& \# Token & $1$ & $2^3$ & $4^3$ & $8^3$ & $64^3$ \\
\midrule
\multirow{4}{*}{\begin{tabular}[c]{@{}l@{}}Stage 3\\ 128$\times$32$\times$32$\times$32\end{tabular}}
& GFLOPs & -- &	$4.3\times10^{-3}$ & $6.4\times10^{-3}$ & $8.21\times10^{-2}$ & $2.84\times10^{2}$ \\
& Time (s)  &-- &  $2.23\times10^{-7}$ & $3.26\times10^{-7}$  & $4.21\times10^{-6}$ & $1.45\times10^{-2}$\\
& Memory (GB) &-- &  $4\times10^{-5}$ & $8.35\times10^{-5}$ & $4.31\times10^{-4}$ & $2.54\times10^{-2}$ \\
& \# Token & $1$  & $2^3$ & $4^3$ & $8^3$ & $32^3$ \\
\midrule
\multirow{4}{*}{\begin{tabular}[c]{@{}l@{}}Stage 4\\ 256$\times$16$\times$16$\times$16\end{tabular}}
& GFLOPs &-- &  $1.3\times10^{-3}$ & $4.3\times10^{-3}$ & $8.76\times10^{-2}$ & 4.56 \\
& Time (s)  &-- & $6.9\times10^{-8}$ & $2.20\times10^{-7}$  & $4.49\times10^{-6}$  & $2.34\times10^{-4}$ \\
& Memory (GB) &-- & $7.79\times10^{-5}$ & $1.50\times10^{-4}$& $7.27\times10^{-4}$& $5.34\times10^{-3}$ \\
& \# Token & $1$ & $2^3$ & $4^3$ & $8^3$ & $16^3$ \\
\midrule
\multirow{4}{*}{\begin{tabular}[c]{@{}l@{}}Stage 5\\ 512$\times$8$\times$8$\times$8\end{tabular}}
& GFLOPs &-- & $8\times10^{-4}$ & $5.7\times10^{-3}$& $1.04\times10^{-1}$& $1.03\times10^{-1}$\\
& Time (s)  &-- & $4.2\times10^{-8}$ & $2.91\times10^{-7}$  & $5.33\times10^{-6}$ & $5.32\times10^{-6}$ \\
& Memory (GB) & -- & $1.54\times10^{-4}$ & $2.83\times10^{-4}$& $1.32\times10^{-3}$& $1.32\times10^{-3}$ \\
& \# Token & $1$ & $2^3$ & $4^3$ & $8^3$ & $8^3$ \\
\midrule
\multirow{3}{*}{Total}
& GFLOPs & --	& $1.80\times10^{-1}$&$2.01\times10^{-1}$& $9.01\times10^{-1}$& $2.36\times10^{6}$ \\
& Time (s)  &--	& $9.24\times10^{-6}$ & $1.03\times10^{-5}$ & $4.62\times10^{-5}$ & $1.21\times10^{2}$\\
& Memory (GB) & --	& $4.55\times10^{-4}$  & $9.18\times10^{-4}$ & $4.74\times10^{-3}$ & 2.55 \\
\midrule
\multicolumn{2}{l}{\textbf{Segmentation (Dice ↑)}}  & 86.15 &	87.11 &	87.72 &	87.74 &	--\\
\multicolumn{2}{l}{\textbf{Transformation (PSNR ↑)}}  & 25.05 &	25.38 &	25.58 & 25.39 &	--\\
\multicolumn{2}{l}{\textbf{Enhancement (PSNR ↑)}}  & 28.81 &	29.42 &	29.66 &	29.76 &	--\\
\bottomrule
\end{tabular}
\end{table*}

\begin{figure*}[htbp]
\centering
\includegraphics[width=0.92\textwidth]{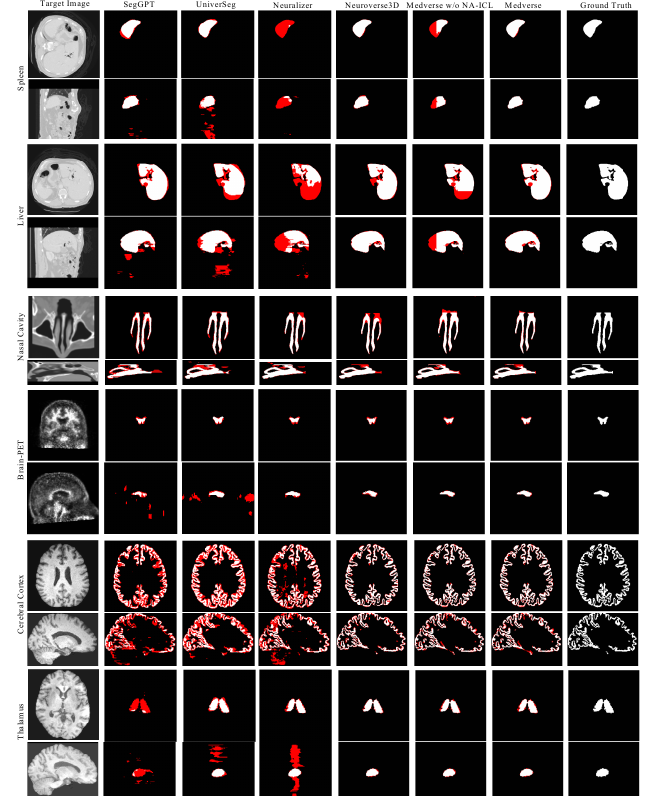}
\caption{Additional quantitative results for segmentation tasks.}
\label{fig:sub_quan1}
\end{figure*}

\begin{figure*}[htbp]
\centering
\includegraphics[width=0.90\textwidth]{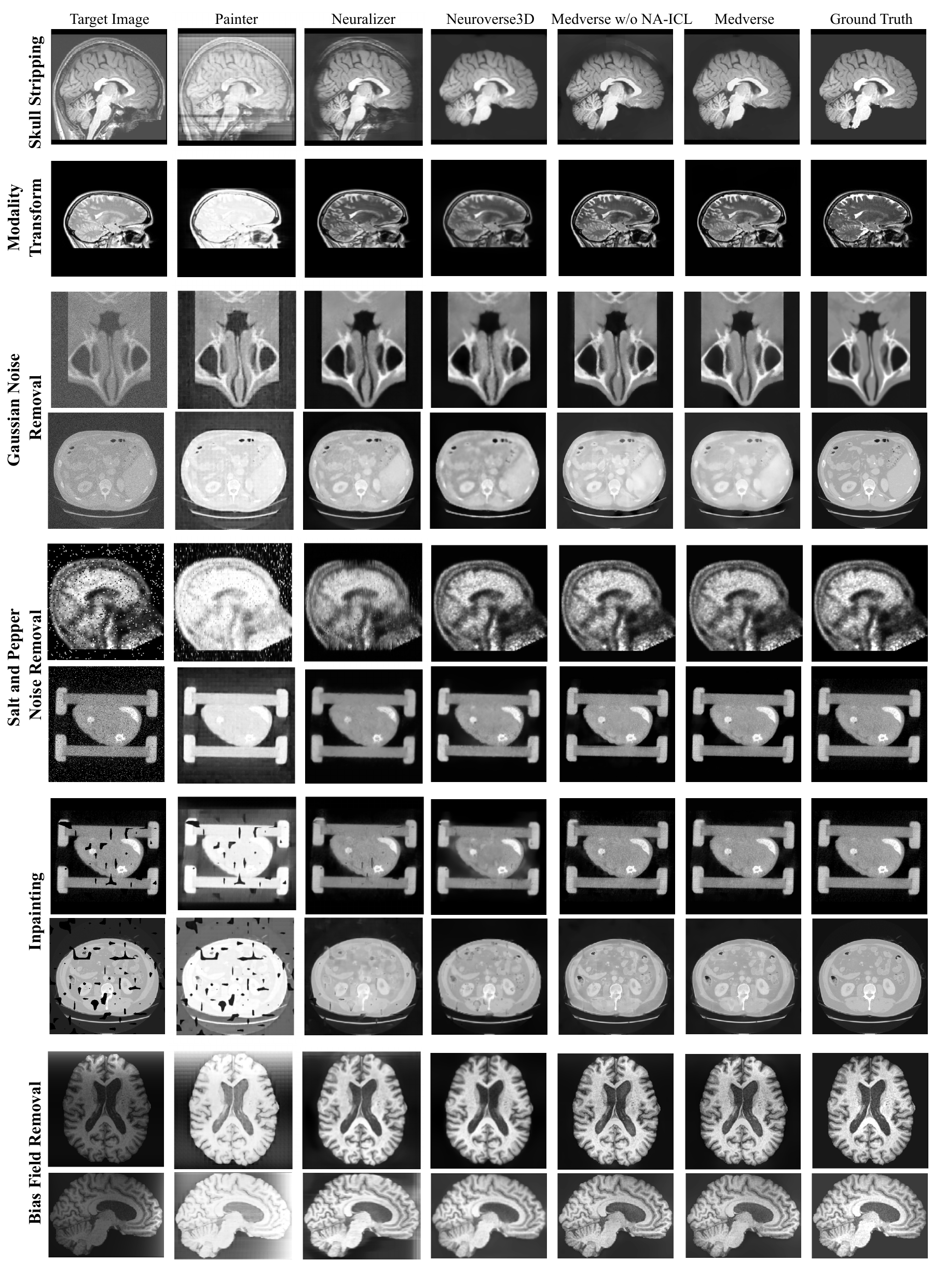}
\caption{Additional quantitative results for transformation and enhancement tasks.}
\label{fig:sub_quan2}
\end{figure*}

\vspace{0.4em}
\noindent\textbf{Additional Quantitative Results.} Figures~\ref{fig:sub_quan1} and~\ref{fig:sub_quan2} present more quantitative results.

\vspace{0.4em}
\noindent\textbf{More Segmentation Results.} Table~\ref{tab:more_seg_comparison} presents the performance of various ICL models on the validation set. It can be observed that our model achieves competitive results on BraTs whole tumor segmentation and vascular segmentation, and significantly outperforms other models in terms of average performance.

\begin{table*}[htbp]
\centering
\setlength{\tabcolsep}{5pt}
\begin{tabular}{l l r r r r r}
\toprule
\textbf{Target} & \textbf{Organ \& Modality} & \textbf{UniverSeg} & \textbf{SegGPT} & \textbf{Neuralizer} & \textbf{Neuroverse3D} & \textbf{Medverse} \\
\midrule
BraTs Whole Tumor            & Brain MRI T1        & 18.13 & 18.39 &  7.32 & \textbf{81.85} & \underline{81.39} \\
BraTs Whole Tumor           & Brain MRI T2       & 34.63 & 30.77 &  6.12 & \underline{89.35} & \textbf{89.59} \\
Topcow                      & Brain MRA      & 38.70 & 24.69 & 13.07 & \textbf{82.30} & \underline{78.98} \\
Cerebral Cortex   & Brain MRI T1  & 69.82 & 50.05 & 66.91 & \underline{87.57} & \textbf{88.44} \\
Hippocampus              & Brain MRI T1  & 42.70 & 23.06 & 60.28 & \textbf{82.14} & \underline{81.42} \\
Thalamus           & Brain MRI T1  & 32.91 & 23.85 & 40.31 & \underline{86.23} & \textbf{86.80} \\
Lateral Ventricle & Brain MRI T1  & 57.90 & 46.13 & 60.02 & \underline{87.41} & \textbf{88.31} \\
Putamen            & Brain MRI T1  & 46.62 & 22.66 & 50.47 & \textbf{85.99} & \underline{85.35} \\
Amygdala           & Brain MRI T1  & 26.56 & 4.53 & 46.38 & \textbf{69.40} & \underline{63.74} \\
PROMISE12                   & Prostate MRI      & \underline{60.50} & 49.93 & 59.15 & 35.02 & \textbf{87.81} \\
MSD Heart                  & Cardiac CT        & \underline{75.18} & 43.94 & 65.12 & 70.30 & \textbf{82.81} \\
RAOS Liver             & Abdominal CT      & 55.81 & 48.54 & 54.54 & \underline{83.71} & \textbf{93.53} \\
RAOS Kidney            & Abdominal CT      & 37.93 & 38.89 & \underline{39.65} & 37.91 & \textbf{86.83} \\
RAOS Stomach           & Abdominal CT      & 25.91 & 22.72 & 16.83 & \underline{32.80} & \textbf{47.69} \\
AMOS22 liver           & Abdominal CT      & 64.94 & 53.33 & 55.71 & \underline{79.09} & \textbf{95.43} \\
AMOS22 Kidney          & Abdominal CT      & 31.92 & 27.89 & 26.23 & \underline{46.80} & \textbf{93.79} \\
AMOS22 Stomach         & Abdominal CT      & \underline{24.64} & 16.16 & 13.42 & 21.68 & \textbf{56.48} \\
\midrule
\multicolumn{2}{c}{\textbf{Average}}                  & 43.81 & 32.09 & 40.09 & 68.21 & \textbf{81.67} \\
\bottomrule
\end{tabular}
\caption{Comparison of segmentation performance on validation set (Dice, \%) across datasets and modalities.}
\label{tab:more_seg_comparison}
\end{table*}

\vspace{0.4em}
\noindent\textbf{Computational Efficiency Comparison.} Table~\ref{tab:model_time} compares the computational efficiency when processing a $128 \times 128 \times 128$ 3D image patch. Medverse exhibits a relatively small parameter size and achieves an inference time comparable to Neuroverse3D, while being significantly faster than other 2D methods. Notably, the inclusion of the autoregressive context introduces only an additional 0.19 seconds of processing time, indicating that the inclusion of the autoregressive context adds minimal computational overhead. For a fair comparison, all 3D methods are evaluated without using sliding-window inference.

\begin{table}[!htpb]
\centering
\setlength{\tabcolsep}{3pt} 
\resizebox{0.7\textwidth}{!}{
\begin{tabular}{lcccc}
\toprule
& Inference Time (s) & Context (pair) & Parameters (M) \\
\midrule
Medverse & 1.16 & 8 3D & 71.05 \\
Medverse w/o NA-ICL & 0.97 & 8 3D & 71.05 \\
Neuroverse3D~\cite{hu2025building} & 1.01 & 8 3D & 70.85 \\
Neuralizer~\cite{czolbe2023neuralizer} & 4.96 & 32 2D & 1.27 \\
UniverSeg~\cite{butoi2023universeg} & 8.36 & 64 2D & 1.18 \\
Painter~\cite{wang2023images} & 31.35 & 1 2D & 307.72 \\
SegGPT~\cite{wang2023seggpt} & 184.89 & 8 2D & 307.72 \\
\bottomrule
\end{tabular}}
\caption{Inference time for a single $128 \times 128 \times 128$ 3D image patch on a V100 GPU along with corresponding model configurations. For Medverse, this includes the processing of both semantic and autoregressive contexts, whereas Medverse w/o NA-ICL excludes the autoregressive context. For 2D comparison methods, we adopt the optimal context settings reported in their respective papers and process the 3D image by splitting it into 128 individual 2D slices.}
\label{tab:model_time}
\end{table}




\end{document}